# Combining Ontology Development Methodologies and Semantic Web Platforms for E-government Domain Ontology Development

Jean Vincent Fonou-Dombeu[1, 2] and Magda Huisman[2]

[1]Department of Software Studies, Vaal University of Technology, South Africa
fonoudombeu@gmail.com
[2]School of Computer, Statistical and Mathematical Sciences, North-West University, South Africa
Magda.Huisman@nwu.ac.za

***ABSTRACT***

*One of the key challenges in electronic government (e-government) is the development of systems that can be easily integrated and interoperated to provide seamless services delivery to citizens. In recent years, Semantic Web technologies based on ontology have emerged as promising solutions to the above engineering problems. However, current research practicing semantic development in e-government does not focus on the application of available methodologies and platforms for developing government domain ontologies. Furthermore, only a few of these researches provide detailed guidelines for developing semantic ontology models from a government service domain. This research presents a case study combining an ontology building methodology and two state-of-the-art Semantic Web platforms namely Protégé and Java Jena ontology API for semantic ontology development in e-government. Firstly, a framework adopted from the Uschold and King ontology building methodology is employed to build a domain ontology describing the semantic content of a government service domain. Thereafter, UML is used to semi-formally represent the domain ontology. Finally, Protégé and Jena API are employed to create the Web Ontology Language (OWL) and Resource Description Framework (RDF) representations of the domain ontology respectively to enable its computer processing. The study aims at: (1) providing e-government developers, particularly those from the developing world with detailed guidelines for practicing semantic content development in their e-government projects and (2), strengthening the adoption of semantic technologies in e-government. The study would also be of interest to novice Semantic Web developers who might used it as a starting point for further investigations.*

## 1. INTRODUCTION

In the past ten years, e-government has been a subject of interest of governments around the world. Governments worldwide are expecting e-government to improve their internal processes and provide Internet and ICT-based service delivery to citizens, businesses and organizations. This requires the design, implementation and launch of web-based systems that present government structures and services online, provide mechanisms for online interaction of government with citizens, and facilitate online citizen participation to government processes and decision making. These mandates of e-government can only be achieved if a large range of government's services and processes are delivered seamlessly to citizens and stakeholders through a single web portal [1], [2]. This raises the issue of developing heterogonous web-based e-government systems of government departments and agencies that can interoperate and be easily integrated. Although the state-of-the-art software engineering techniques including object-oriented and agile methods provide appropriate solutions to the aforementioned engineering problems of services integration and interoperability in e-government [1], [3], [4], it has been demonstrated that they have certain limitations [3], [4], [5], [6]. Therefore, during the past six years, Semantic Web technologies have emerged as promising solutions to these problems [3], [4], [7], [8], [16].

Semantic-based e-government consists of describing existing entities, concepts, processes, laws and regulations governing a government service domain, into a conceptual model namely domain ontology. This domain ontology is initially represented in a human readable form. Then, to make it processable by computers, ontology editing and implementing platforms should be used by e-government developers to develop the domain ontology in Semantic Web machine processable syntaxes such as XML, RDF, and OWL. In light of the above, practicing semantic development in e-government could be challenging to e-government system developers without any knowledge of ontology and Semantic Web technologies. Thus, there is a need for detailed research in the field of e-government that focuses on ontology development using available ontology development methodologies and open-source Semantic Web development platforms. To the best of our knowledge, current research focusing on semantic e-government development [3], [7], [8], [9], [10], [11], [12], [13], [14], [15], [16] has not yet filled this gap. Furthermore, only a few of current works provide detailed guidelines for developing government domain ontology from a government service domain.

This research presents a case study combining an ontology building methodology and two state-of-the-art Semantic Web platforms, namely Protégé and Java Jena ontology API, for semantic ontology development in e-government. Firstly, a framework adopted from the Uschold and King [17] ontology building methodology is employed to build a domain ontology describing the semantic content of a government service domain. Thereafter, UML is used to semi-formally represent the domain ontology. Finally, Protégé and Jena API are employed to create the Web Ontology Language (OWL) and Resource Description Framework (RDF) representations of the domain ontology respectively to enable its computer processing. The study aims at: (1) providing e-government developers, particularly those from the developing world, where there is little or no use of Semantic Web technologies in e-government, with detailed guidelines for practicing semantic content development in their e-government projects and (2), strengthening the adoption of semantic technologies in e-government. The study would also be of interest to novice Semantic Web developers who might use it as a starting point for further investigations.

The rest of the paper is organized as follows. Section 2 presents existing languages for ontology representation in Semantic Web. The software platforms for ontology development are described in Section 3. Section 4 conducts a literature review on the state of usage of Semantic Web technologies in e-government to date. The existing methodologies for building ontologies as well as the stages of ontology development are discussed in Section 5. Section 6 presents a case study development of a government domain ontology combining ontology development methodology and Semantic web platforms, including Protégé and Jena API. A discussion is carried out in Section 7 and a conclusion is drawn in the last section.

## 2. ONTOLOGY LANGUAGES FOR THE SEMANTIC WEB

The Semantic Web [18] is an evolution of the current Web that provides meanings to Web contents to enable their intelligent processing by computers. The meanings of Web contents are represented with ontology and described formally in logic-based syntaxes to facilitate their integration and accessibility over the Web [19]. The logic-based description of ontology is carried out with Semantic Web ontology languages such as XML, XML Schema (XMLS), RDF, RDF Schema (RDFS), and OWL. The work presented in this paper uses RDF(S) and OWL which are the leading Semantic Web ontology languages [6], [19], [20], [21].

### 2.1 RDF(S)

RDF and RDFS are the first standardized Web Based languages [19], [20]. RDF is a data model used to describe resources on the Web, whereas, RDFS is an improved version of RDF which provides facilities for the definition of basic ontology elements such as classes and hierarchy of classes, properties, domain and range of properties [19], [20]. RDF uses statements in the form of *<S, P, O>* to represent an ontology. The meaning of a RDF statement is that subject S has property P with value O. In a RDF statement, S and P are uniform resource identifiers (URIs), while O is either a URI or a literal value [6].

### 2.2 OWL

OWL was developed to overcome the weak expressive power of RDF(S) [19], [20], [21]. The expressivity of RDF(S) is enhanced by OWL with tools for: describing relations between classes, defining properties' characteristics, cardinality and value restrictions on properties, etc. [19], [20].

In practice, Semantic Web developers do not have to write RDF(S) and/or OWL codes by hand; several software platforms exist for the automatic development of RDF(S) and OWL codes. The next section expands more on existing software platforms for Semantic Web ontology development.

## 3. SOFTWARE PLATFORMS FOR SEMANTIC WEB ONTOLOGY DEVELOPMENT

To support the vision of the Semantic Web which is making machine-readable content available on the Web, several software platforms and application interfaces (APIs) have been developed to permit the automatic creation and use of RDF(S) and OWL ontologies. Recent comparative studies of some of the existing RDF(S) and OWL editing platforms are provided in [21], [22]. A more exhaustive list of these platforms could be found in [21], [22], [23], [27]; they include Protégé, WebODE, OntoEdit, KAON1, and so forth. Beside the software platforms used for the edition of RDF(S) and OWL ontologies, there exist APIs such as OWL API [23], Jena API [6], Sesame [5], etc., which provide facilities for the persistence storage and query of RDF(S) and OWL ontologies. Protégé and Jena API are discussed and used in this study as they are the leading platforms for Semantic Web development [6], [21], [22]; furthermore, they are both open-source software and might facilitate the repeatability of this study.

### 3.1 Protégé

Protégé is an open-source platform developed at Stanford Medical Informatics. It provides an internal structure called model [23] for ontologies representation and an interface for the display and manipulation of the underlying model. The Protégé model is used to represent ontology elements as classes, properties or slots, property characteristics such as facets and constraints, and instances. The Protégé graphical user interface can be used to create classes and instances, and set class properties and restrictions on property facets. Additionally, Protégé has a library of various tabs for the access, graphical visualization, and query of ontologies. Protégé can be currently used to load, edit and save ontologies in different formats including XML, RDF, UML, and OWL [23].

### 3.2 Jena API

Jena is a Java ontology API. It provides object classes for creating and manipulating RDF graphs called interfaces. A RDF graph is called a model and represented with the Model interface. The resources, properties and literals describing RDF statements are represented with the Resource, Property and Literal interfaces respectively. Jena also provides methods that allow saving and retrieving RDF graphs to and from files. The Jena platform supports various database management systems such as PostgreSQL, MySQL, Oracle, and so on; it also provides various tools including RDQL query language, a parser for RDF/XML, I/O modules for RDF/XML output, etc. [6]. The state of use of Semantic Web technologies in e-government is presented in the next section.

## 4. BACKGROUND ON THE USE OF SEMANTIC WEB TECHNOLOGIES IN E-GOVERNMENT

In the past six years, e-government has been one of the most active areas of Semantic Web development. However, current research practicing semantic development in e-government does not focus on the application of available methodologies and platforms for developing government domain ontologies. In [9], a process-document ontology describing the document hierarchy and the structure of the business processes in e-government is modelled using RDF and OWL graphical representations. However, no practical implementation was discussed to illustrate how the proposed process-document ontology can be created with existing Semantic Web editing and representation platforms.

Xiao et al. [8] and Sabucedo et al. [16] proposed specific e-government domain ontologies to address the issue of service integration in e-government; the proposed ontology models were represented in OWL for semantic processing. However, no information was given on the platforms employed to generate the OWL versions of the proposed ontology models.

Other semantic-based solutions for e-government services integration and interoperability based on specific ontology models are proposed in [7], [10]. The proposed ontology models were developed in OWL using dedicated platforms namely ONTOGOV [7] and IESD [10], respectively. However, the platforms employed are proprietary and little information is provided for their widespread use within the e-government community.

The analysis of other relevant literature by Sabucedo and Rifon [11], Chen et al. [12] and Gugliotta et al. [13] reveals other specific ontology models describing various aspects of e-government service delivery. However, the proposed ontology models were only presented at the conceptual level. Further, none of the works has provided clues on how the proposed ontology models can be constructed from the complex public administration system with existing ontology building methodologies.

The issue of services interoperability in e-government is further addressed in Muthaiyah and Kerschberg [3], Zhang and Wang [14] and Bettahar et al. [15] with ontology-based solutions; the ontology components of the proposed solutions are modeled in OWL [3], [14], [15] and Sematic Web Rule Language (SWRL) [3], [14] with Protégé to enable the semantic interoperability of e-government services. Although these studies have used Protégé for semantic ontology development, they did not focus on the methodological approach for semantic ontology development in e-government; then, it is not clear how the ontology components of the proposed interoperability solutions were built from the complex public administration system until their implementation with Protégé.

In summary, the above research demonstrate the interest in Semantic Web technologies based on ontology in e-government; they show how various ontology models are being used in e-government researches and projects to describe and specify e-government services, aiming at their semantic integration and interoperability. However, the ontology models employed were only presented at the conceptual level [9], [11], [12], [13]; in certain cases, no details of the implementation platforms were given [8], [16]; and some of them were developed with proprietary platforms [7], [10]. Further, none of the current research provides guidelines for constructing the proposed semantic ontology models from the complex public administration system using existing ontology building methodology provided in the ontology engineering field. All these factors do not facilitate the repeatability of various ontology-based solutions for e-government services integration and interoperability that are being proposed in various e-government researches and projects nor do they contribute to the widespread adoption of Semantic Web technologies in e-government. The next section explains the ontology development process.

# 5. ONTOLOGY DEVELOPMENT PROCESS

## 5.1 Ontology Development Methodologies

At present, there is no universal definition of ontology in the literature. Ontology is commonly defined as an explicit specification of a conceptualization [24] i.e., a model of the real world domain such as medicine, geographic information system, physics, e-government and so forth; which is explicitly represented with existing objects, concepts, entities and relationships between them. The above definition of ontology shows that developing an ontology for a given domain of knowledge could be a complex and challenging task [23]; especially for a complex domain such as the public administration system. Therefore, ontology developers need appropriate methodologies that will guide them in the process of building an ontology characterizing a particular domain.

To this end, a considerable amount of research has been done in the field of ontology engineering to develop proper methodologies for ontology development. Detailed comparative studies of these methodologies are provided in [25], [26], [27]. The ontology development methodologies prescribe various steps and tasks that must be performed when building ontology. To the best of our knowledge, none of the current research practicing semantic ontology development in e-government has referred to a particular methodology that was employed when building the proposed e-government specific ontology models. This may not ease the repeatability of the proposed ontology-based solutions by e-government developers without any knowledge of ontology and Semantic Web technologies.

The primary objective of this study is to provide e-government developers, particularly those from the developing world, with guidelines for practicing Semantic content development in their e-government projects. Therefore this research presents a case study of semantic ontology development in e-government. A framework adopted from the Uschold and Kind [17] ontology building methodology is applied to build a domain ontology from a government service domain.

The domain ontology is further developed in RDF and OWL with Protégé and Java Jena ontology API to enable its semantic processing by computers.

### 5.2 Phases of Ontology Development

A rigorous development process for ontology building requires the use of its development methodologies and platforms. A methodology mainly prescribes guidelines for the specification, conceptualization, formalization and implementation of ontology [27]. The specification phase defines the aims and roles of the intended ontology as well as people who will be using it. During the conceptualization phase, a conceptual or domain ontology is built. In its simple representation, a conceptual or domain ontology is a graph where the vertices are objects, concepts, and entities of the domain, and the edges are lines interconnecting pairs of vertices and representing the relationships between the constituents of the domain. The formalization phase transforms the conceptual model or domain ontology into its semi-formal representation; this can be done either in description logic [29] or UML formalisms [30]. The UML formalism is used in this study because it is widely used by software developers for object-oriented systems development. During the implementation phase of ontology development, the semi-formal version of the ontology is formally represented in one of Semantic Web languages with ontology editing platforms (see sections 2 and 3).

After its formal representation, the developed ontology has to be effectively used by its intended users. Therefore, the formal ontology must be deployed onto the Web using programming platforms such as Java, C++, .NET, etc. This provides facilities for database storage and querying of the developed ontology.

To fulfil the objectives of this study, which are to provide e-government developers with detailed guidelines for practicing semantic content development in their e-government projects and strengthen the adoption of semantic technologies in e-government, the main phases of ontology development presented above are applied to develop a government domain ontology using a framework adopted from the Uschold and King [17], [28] methodology and Protégé and Java Jena ontology API platforms.

The Uschold and Kind [17] methodology was chosen in this study for its clarity and the fact that it is technology and platform independent [25]; which might facilitate its use by novice ontology developers and promote a fast development of domain ontologies.

## 6. APPLICATION

This section presents a case study development of government domain ontology.

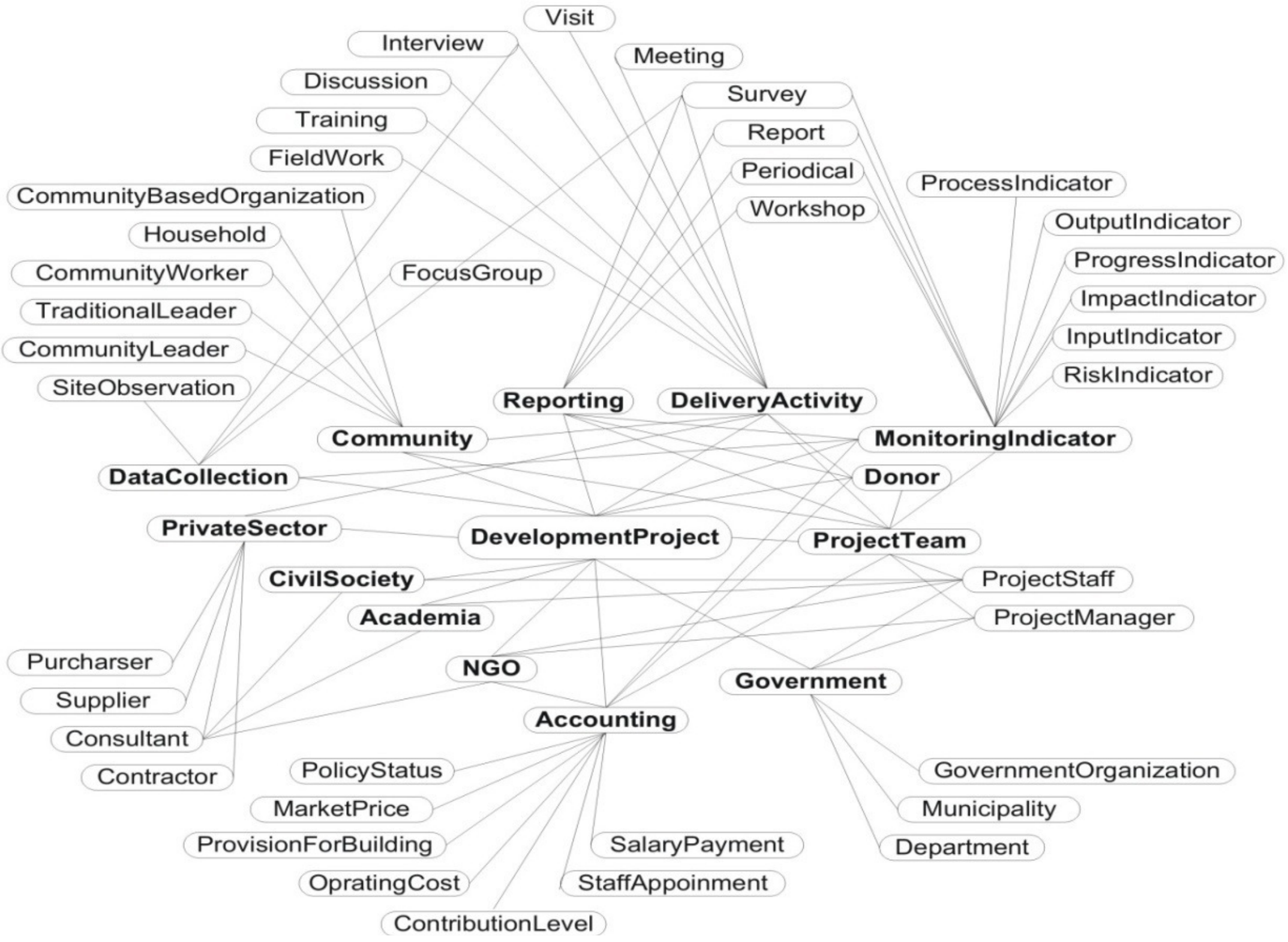

Figure 1: OntoDPM Domain Ontology

## 6.1 Presentation of the Government Service Domain

The government service domain used in this study is the domain of development projects monitoring in developing countries. In fact, in developing countries as well as in Sub Saharan Africa (SSA), almost every government department is somehow involved in the implementation of a programme aiming at improving the welfare of its people. These programmes are commonly called development projects and include infrastructure development, water supply and sanitation, education, rural development, health care, ICT infrastructure development and so forth. Thus, application that could interface all the activities related to development projects implementation in a SSA country could bring tremendous advantages. For example, such a web application would improve the monitoring and evaluation of projects and provide transparency, efficiency and better delivery to populations. In a previous study [28], we built a domain ontology of development projects monitoring (OntoDPM) for such an application. The next subsections describe the methodology and platforms employed.

## 6.2 OWL Representation of the OntoDPM Domain Ontology

A framework adopted from the Uschold and King [17] ontology building methodology was used in [28] to build the OntoDPM as in Figure 1. The OntoDPM shows the key concepts of the domain (people, stakeholder, financier, monitoring indicator, reporting technique, etc.), the activities carried out in the domain (training, discussion, fieldwork, visit, meeting, etc.) and the relationships between the constituents of the domain. This study does not expand on the framework for building the OntoDPM in Figure 1. Interested readers may refer to [28] for detailed information. The OntoDPM in Figure 1 was further written semi-formally in UML and implemented in OWL using Protégé [31], [32].

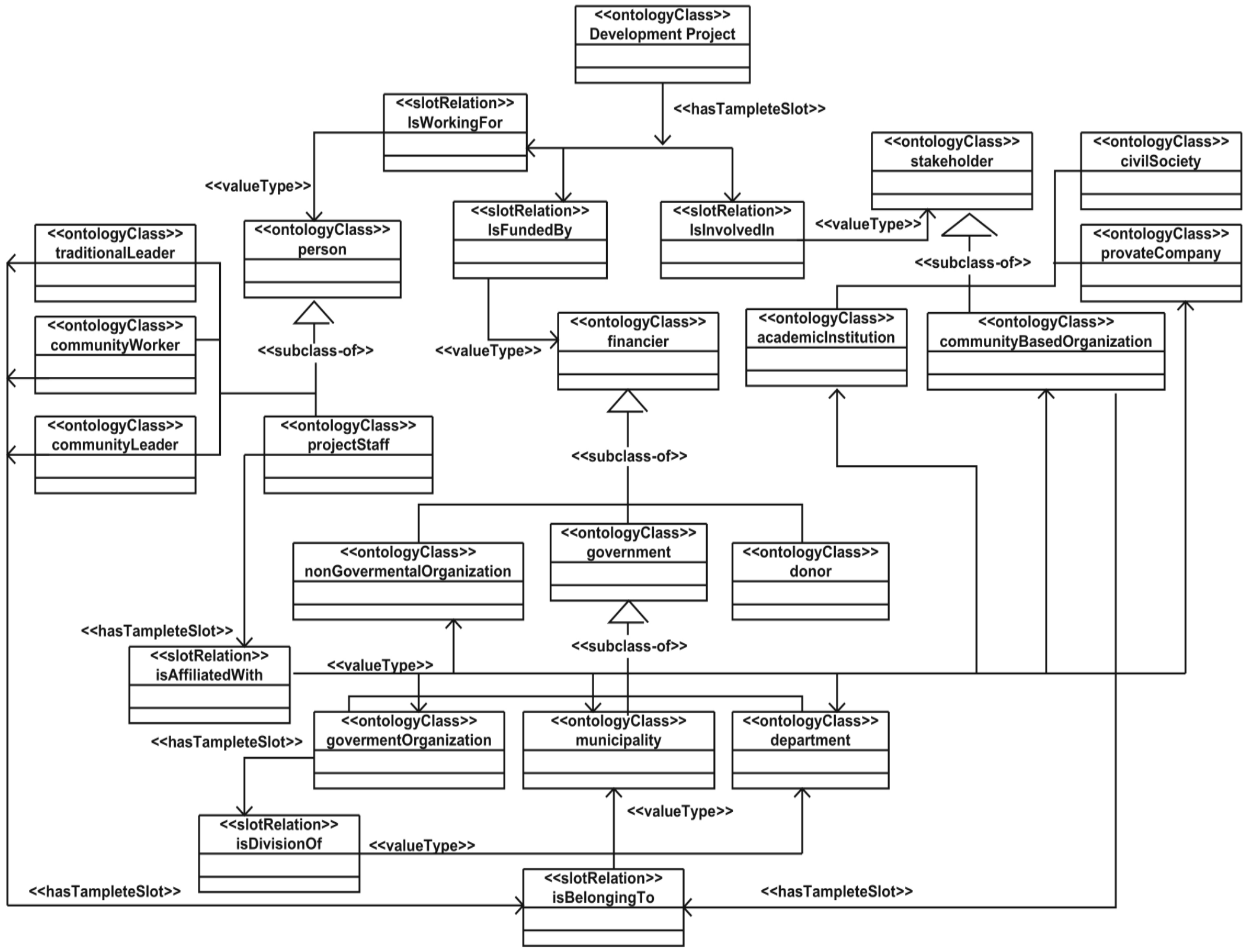

Figure 2: Part of UML Class Diagram of the OntoDPM Domain Ontology

The UML representation of the OntoDPM was constructed by identifying classes and class instances in the OntoDPM in Figure 1 and categorizing relationships between classes (composition, association, inheritance). Figure 2 represents a part of classes, inheritance structure and properties/slots of the OntoDPM, in the UML formalism for knowledge representation [30]. This formalism was chosen because it allows modeling ontologies with instances/individuals, slots and classes, which are the terminologies of Protégé as well [31]. In this formalism, a class is labelled "*ontology class*", a property/slot "*slot relation*" and an instance/individual "*IndividualOf*". Properties/slots represent the relationships between the concepts of the domain ontology. Each slot has a domain and a range which are labelled *hasTempletateSlot* and *valueType,* respectively. The inheritance relationship between classes is labelled "*subclass-of*".

The OWL version of the OntoDPM was created with Protégé (see part in Table 1) using its semi-formal representation (see part in Figure 2). Firstly, we downloaded Protégé version 4.0 from the Stanford Medical Informatics website and installed it in our computer; related documentations [31], [32] were downloaded as well. Thereafter, the user-friendly graphical user interface of Protégé was used to create the class hierarchy (classes, instances and inheritance structure), slots, and domain and range of slots respectively. These features of the OntoDPM were created based on its UML representation (see part in Figure 2). The resulting Protégé file was saved as an OWL file onto the disc.

Table 1: Part of OWL Version of the OntoDPM Domain Ontology

```
---
<owl:Class rdf:about="#Stakeholder">
<owl:Class rdf:about="#CivilSociety">
<rdfs:subClassOf rdf:resource="#Stakeholder"/>
</owl:Class>
 <owl:Class rdf:about="#PrivateCompany">
 <rdfs:subClassOf rdf:resource="#Stakeholder"/>
 </owl:Class>
 <owl:Class rdf:about="#AcademicInstitution">
 <rdfs:subClassOf rdf:resource="#Stakeholder"/>
 </owl:Class>
 <owl:Class rdf:about="#CommunityBasedOrganization">
 <rdfs:subClassOf rdf:resource="#Stakeholder"/>
 </owl:Class>
 <owl:Class rdf:about="#Financier">
 <owl:Class rdf:about="#Donor">
 <rdfs:subClassOf rdf:resource="#Financier"/>
 </owl:Class>
 <owl:Class rdf:about="#Government">
 <rdfs:subClassOf rdf:resource="#Financier"/>
 </owl:Class>
 <owl:Class rdf:about="#NonGovernmentalOrganization">
 <rdfs:subClassOf rdf:resource="#Financier"/>
 </owl:Class>
   ---
<owl:ObjectProperty rdf:about="#IsFunderBy">
<rdfs:domain rdf:resource="#DevelopmentProject"/>
 <rdfs:range rdf:resource="#Financier"/>
</owl:ObjectProperty>
<owl:ObjectProperty rdf:about="# IsInvolvedIn ">
<rdfs:domain rdf:resource="#DevelopmentProject"/>
<rdfs:range rdf:resource="#Stakeholder"/>
</owl:ObjectProperty>
---
```

The code in Table 1 was obtained by opening the saved OWL file with open-source programming editors such as JCreator and JGrasp. Table 1 presents parts of the *subclass-of* relationships (inheritance) and domains and ranges of slots in the OntoDPM. It is worth noting that the concepts of class, property/slot, domain and range, and individual/instance used in the UML representation of the OntoDPM (see part in Figure 2), are concepts of Protégé as well; this indicates that the formalism employed in Figure 2 is appropriate for representing a semi-formal ontology to be implemented with Protégé. A complete description of OWL syntax is beyond the scope of this research; the study aims to provide guidelines for creating OWL ontology with Protégé, based a semi-formal UML representation of domain ontology such as the OntoDPM. Further information on the OWL syntax and its development with Protégé could be found in [31] and [32]. In the following subsection, Java Jena ontology API is employed to create the RDF formal representation of the OntoDPM.

## 6.3 RDF Representation of the OntoDPM Domain Ontology

Let’s recall that RDF is a Semantic Web language for representing resources on the Web. A resource is any information that can be accessed over the Web using an URI. Three concepts are fundamental in RDF including subject (S), predicate (P) and Object (O). A subject is a resource to be referred to on the Web; a predicate is a property describing the resource and an object is a value of the property. Thus, the triplet <S, P, O> describes the same information and forms an RDF graph or statement [6]. An RDF graph or statement is graphically represented with two nodes and one arc; where, the arc is the property that describes the resource and the nodes at both sides of the arc, the resource and property’s value respectively. In light of the above, the RDF syntax represents each class of an ontology as a resource which has properties with values.  Thus, an ontology will be represented in RDF with several statements where each statement or a set of related statements forms an RDF sub-graph of the entire RDF graph of the intended ontology.

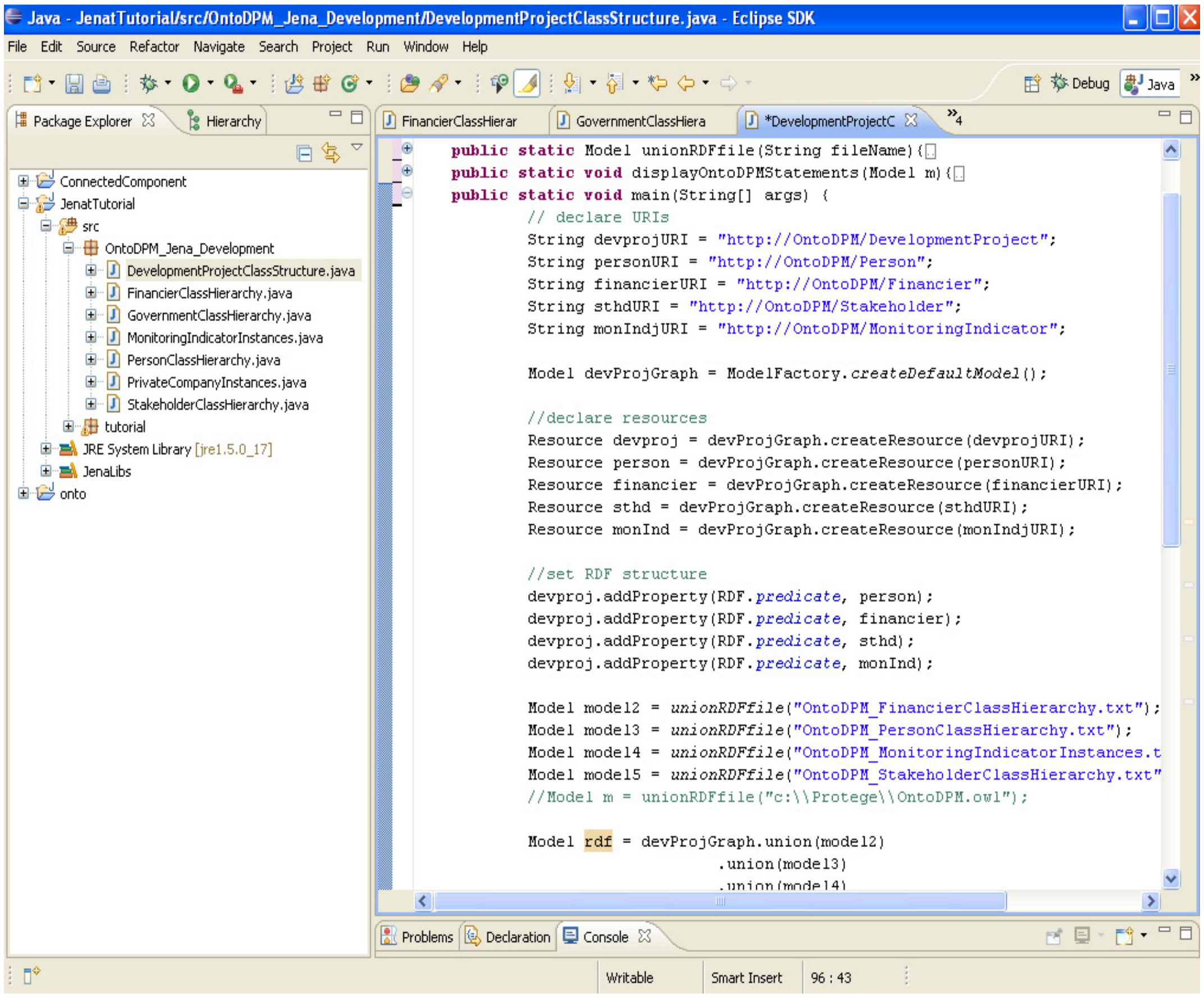

Figure 3: Screenshot of the Jena Implementation of the OntoDPM Domain Ontology

The RDF version of the OntoDPM was created with the Java Jena Ontology API. Firstly, we downloaded, installed and configured the Jena API in the Eclipse Java software development kit (SDK). As there are many classes in the OntoDPM, it was difficult to handle the same Jena RDF model to create all of them. We found that the Jena Model interface provides the union method that can be used to integrate different branches of a large RDF graph. Then, we created the main components (inheritance and instances) of the OntoDPM as individual Java files (see the left part of Figure 3). In each of the Java files, the main() method was included to generate and write the corresponding RDF sub-graph into a text file. Thereafter, we coded a method to read each text file and construct its RDF sub-graphs with the Jena Model interface. Finally, the individual RDF sub-graphs were integrated with the union method of the Model interface, in a unique RDF graph of the OntoDPM.

Table 2: Part of RDF Representation of the OntoDPM Domain Ontology

```
<rdf:RDF
  xmlns:rdf=http://www.w3.org/1999/02/22-rdf-syntax-ns#>
 <rdf:Description rdf:about="http://OntoDPM/DevelopmentProject">
 <rdf:predicate>
  <rdf:Description rdf:about="http://OntoDPM/Stakeholder">
   <rdf:predicate rdf:resource="http://OntoDPM/CivilSociety"/>
   <rdf:predicate>
    <rdf:Description rdf:about="http://OntoDPM/PrivateCompany">
     <rdf:object rdf:resource="http://OntoDPM/Consultant"/>
     <rdf:object rdf:resource="http://OntoDPM/Purchaser"/>
     <rdf:object rdf:resource="http://OntoDPM/Contractor"/>
     <rdf:object rdf:resource="http://OntoDPM/Supplier"/>
    </rdf:Description>
   </rdf:predicate>
   <rdf:predicate rdf:resource="http://OntoDPM/AcademicInstitution"/>
   <rdf:predicate rdf:resource="http://OntoDPM/CommunityBasedOrganization"/>
  </rdf:Description>
 </rdf:predicate>
  <rdf:Description rdf:about="http://OntoDPM/Financier">
   <rdf:predicate rdf:resource="http://OntoDPM/NonGovernmentalOrganization"/>
   <rdf:predicate>
    <rdf:Description rdf:about="http://OntoDPM/Government">
     <rdf:predicate rdf:resource="http://OntoDPM/Agency"/>
     <rdf:predicate rdf:resource="http://OntoDPM/Municipality"/>
     <rdf:predicate rdf:resource="http://OntoDPM/Department"/>
     <rdf:object>Institute of Nuclear Research</rdf:object>
     <rdf:object>Department of Health</rdf:object>
     <rdf:object>Ekhurhuleni</rdf:object>
    </rdf:Description>
   </rdf:predicate>
   <rdf:predicate rdf:resource="http://OntoDPM/Donor"/>
  </rdf:Description>
 </rdf:predicate>
---
```

```
 </rdf:Description>
</rdf:RDF>
```

A part of the code for integrating the RDF sub-graphs is depicted on the right side of Figure 3. Table 2 shows a part of the generated RDF code of the OntoDPM. One can notice that RDF code is not as readable as an OWL code because of the weak expressive power of RDF mentioned earlier in this study. However, Table 2 illustrates the representation of a part of the OntoDPM class hierarchy and instances with basic RDF concepts of resource, predicate and object. A complete explanation of the development of RDF ontology with Jena API is beyond the scope of this research. Further information on how to use Jena API to develop ontology in RDF syntax could be found in [33], [34]. A discussion and future direction of our research are presented in the next section.

# 7. DISCUSSION

The study has discussed Semantic Web ontology languages and software platforms for ontology development. A domain ontology was built with a framework adopted from the Uschold and King [17] ontology building methodology [28]. A semi-formal representation of the domain ontology was done with the UML formalism. Further, two state-of-the-art Semantic Web platforms for ontology development including Protégé and Java Jena API were used to generate the machine processable version of the domain ontology in OWL and RDF, respectively. It is worth highlighting that Jena API provides some useful features for ontology deployment. In fact, the Jena API provides the RDQL language for RDF storage and query with various database management systems including PostgreSQL, MySQL, Oracle [6], etc. However, the database storage and query of RDF ontology is out of the scope of this research and will be the focus of our future work. Furthermore, Jena API provides parsing mechanisms that could be exploited to read OWL ontology developed with Protégé and generate an RDF graph; this might facilitate the development of real-world Semantic Web applications where ontology edition is done with Protégé, while queries are handled with Jena interfaces. The parsing mechanism for bridging Protégé and Jena API will be investigated in our future work as well.

# 8. CONCLUSION

The study reviews the literature and points out that current research focusing on Semantic Web development in e-government does not refer to any existing ontology development methodology when reporting on either the e-government specific ontology models that they have developed or an ontology-based solution for e-government services integration and interoperability that they propose. The study has also pointed out that current research does not provide detailed guidelines for developing government domain ontology using available open-source Semantic Web platforms. All these factors do not ease the repeatability of the various specific ontology models being developed in e-government domain nor do they strengthen the adoption of semantic technologies in e-government.

As a solution, the study has developed a government domain ontology using a framework adopted from the Uschold and King ontology building methodology and two leading Semantic Web development platforms namely Protégé and Java Jena Ontology API. The framework used for ontology building provides clearly defined steps and their application and it is platform independent as well [28]. This may facilitate its use by novice ontology developers and promote a fast development of domain ontology. Furthermore the Semantic Web platforms employed including Protégé and Java Jena Ontology API are all open-source software which are downloadable from the Internet. This may facilitate the repeatability of the study within the e-government development community and strengthen the adoption of semantic technologies in e-government.

## Authors

[1]**J.V. Fonou Dombeu** is a PhD candidate at the School of Computer, Statistical and Mathematical Sciences at the North-West University, South Africa and a Lecturer in the Department of Software Studies at the Vaal University of Technology, South Africa. He received an MSc. in Computer Science at the University of KwaZulu-Natal, South Africa, in 2008, BSc. Honour's and BSc. in Computer Science at the University of Yaoundé I, Cameroon, in 2002 and 2000 respectively. His research interests include: Biometric for Personal Identification, Ontology, Agent Modelling, and Semantic Knowledge representation in e-Government.

[2]**M. Huisman** is a professor of Computer Science and Information Systems at the North-West University (Potchefstroom Campus) where she teaches software engineering, management information systems, and decision support systems. She received her Ph.D degree in Computer Science and Information Systems at the Potchefstroom University for CHE in 2001. Magda is actively involved in research projects regarding systems development methodologies. Her research has appeared in journals such as *Information & Management* and she has presented papers at international conferences in China, Australia, Switzerland, Canada, Japan and Latvia. Her current research interests are in systems development methodologies and the diffusion of information technologies.